\documentclass[runningheads]{llncs}
\usepackage[T1]{fontenc}
\usepackage{graphicx}
\usepackage{booktabs}
\usepackage{placeins}
\usepackage{hyperref}

\begin{document}
\sloppy
\title{Rad-ReStruct: A Novel VQA Benchmark and Method for Structured Radiology Reporting} 
\titlerunning{Rad-ReStruct}
\author{Chantal Pellegrini\thanks{Contributed equally. Corresponding author: chantal.pellegrini@tum.de} \and
Matthias Keicher\textsuperscript{$\star$} \and
Ege Özsoy \and
Nassir Navab}

\authorrunning{Pellegrini and Keicher et al.}

\institute{Computer Aided Medical Procedures, Technical University Munich, Germany}

\maketitle
\begin{abstract}
Radiology reporting is a crucial part of the communication between radiologists and other medical professionals, but it can be time-consuming and error-prone. One approach to alleviate this is structured reporting, which saves time and enables a more accurate evaluation than free-text reports. However, there is limited research on automating structured reporting, and no public benchmark is available for evaluating and comparing different methods. To close this gap, we introduce Rad-ReStruct, a new benchmark dataset that provides fine-grained, hierarchically ordered annotations in the form of structured reports for X-Ray images. We model the structured reporting task as hierarchical visual question answering (VQA) and propose hi-VQA, a novel method that considers prior context in the form of previously asked questions and answers for populating a structured radiology report. Our experiments show that hi-VQA achieves competitive performance to the state-of-the-art on the medical VQA benchmark VQARad while performing best among methods without domain-specific vision-language pretraining and provides a strong baseline on Rad-ReStruct. Our work represents a significant step towards the automated population of structured radiology reports and provides a valuable first benchmark for future research in this area. Our dataset and code is available at \href{https://github.com/ChantalMP/Rad-ReStruct}{https://github.com/ChantalMP/Rad-ReStruct}.

\end{abstract}
\section{Introduction}
Radiology is a critical medical field that relies on accurate and efficient communication between radiologists and other healthcare professionals enabled through radiology reports. However, generating these reports takes a lot of time and is prone to errors, as it often relies on ambiguous natural language. One alternative to free-text reports is to use structured reporting, which is endorsed by radiology societies, saves time, and offers standardized content and terminology \cite{nobel2022structured,hong2013content}.

Automated report generation can reduce radiologists' workload and support quick diagnostic decisions. Most current research focuses on generating free-text reports, which lack standardization, and still face challenges of ambiguity and difficulties in clinical correctness evaluation \cite{hou2021ratchet,tanwani2022repsnet,wang2022medical,li2022self,pino2021clinically}. In comparison, automating structured reporting allows an accurate evaluation of clinical correctness and can enforce the prediction of detailed findings. However, for automating structured reporting, the research is limited. Some studies predict high-level abnormalities using pre-defined template sentences \cite{pino2021clinically,keicher2022few}, or predict location and attributes for a single disease \cite{bhalodia2021improving}. Syeda-Mahmood et al. \cite{syeda2020chest} predict fine-grained but unstructured labels to retrieve and adapt free-text reports from a database. However, none of these works predict highly-detailed and structured annotations as needed to populate an entire structured report. A significant challenge towards this goal is the lack of public benchmarks with highly detailed structured annotations. To facilitate future research, we introduce Rad-ReStruct, the first dataset of publicly available, fine-grained, and structured annotations for Chest X-Rays. To create Rad-ReStruct, we define a detailed, multi-level structured reporting template and automatically populate it by parsing and analyzing unstructured finding summaries from the IU X-ray dataset \cite{iu-xray}. 

Structured reports with high standardization have a structured layout and content, e.g., organized in expendable trees with drop-down menus to select answers \cite{nobel2022structured}. A user interface for structured reporting would pose a series of questions that, dependent on the answer, lead to expandable follow-up questions. In this setup, structured reporting can be considered several classification tasks on different levels. We model this as a hierarchical visual question answering (VQA) task and propose hi-VQA, a hierarchical, autoregressive VQA method for populating structured reports by successively filling out all fields in the report while preserving consistency. Hi-VQA considers the prior context of previously asked questions and answers, allowing to exploit inter-dependencies between questions about the same image. For structured report population, this is essential, as lower-level questions directly depend on higher levels. Further, our autoregressive formulation enhances explainability, showing at which level and for which question type the model made mistakes. As backbone, we propose a simple yet effective VQA architecture relying on large pretrained image and text encoders and a transformer-based fusion module. Using VQA \cite{antol2015vqa} allows to exploit the knowledge encoded in large language models. It has recently received attention in the medical field, mainly on small datasets, where every question is answered independently \cite{nguyen2019overcoming,ren2020cgmvqa,khare2021mmbert,tanwani2022repsnet,chen2022multi}. One previous work explicitly models question consistency for medical VQA in the loss \cite{tascon2022consistency}. Another work had promising results using an unstructured question history in a visual dialog setting \cite{kovaleva2020towards}. They ask for high-level abnormalities, such as "Pneumonia?" and use a randomly sampled, fixed history of other abnormality questions. In contrast, we define a hierarchical history with detailed questions and an autoregressive model. 

We demonstrate the effectiveness of our streamlined design and hierarchical framework in our experimental results, reaching competitive results to the SOTA on the medical VQA benchmark VQARad and setting a baseline for Rad-ReStruct. Overall, our work is a significant step towards automating the population of structured radiology reports and provides a valuable benchmark for future research in this area.\looseness=-1
\section{Methodology}

\subsection{Rad-ReStruct Dataset}
We propose the first benchmark dataset to enable the development and comparison of methods for the population of structured reports entailing hierarchical and fine-grained classifications for radiological images. Rad-ReStruct is based upon the IU-Xray dataset \cite{iu-xray} and consists of X-Ray images paired with fine-grained radiological findings organized as a structured report. To create the dataset, we first define a detailed report template and then populate it automatically by parsing and analyzing the unstructured expert annotations of the reports in IU-Xray.\looseness=-1
\begin{figure}[tb]
\includegraphics[width=\textwidth]{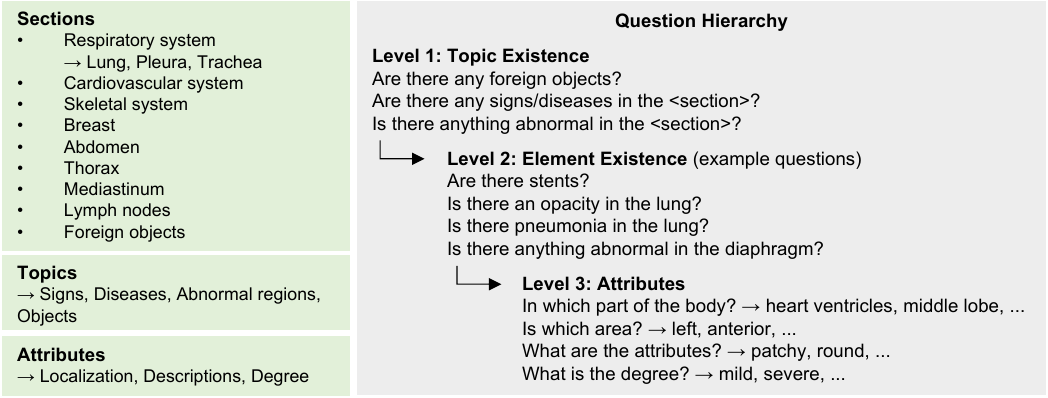}
\caption{Overview of report and question structure.}
\label{fig_template}
\centering
\end{figure}

\textbf{Creation of Structured Report Template} We build upon the semi-structured encoded findings provided for the IU-XRay data collection \cite{iu-xray}. The encoded findings were provided by medical experts, who labeled the IU-XRay free-text reports using MeSH (Medical Subject Headings) \cite{fb1963medical} and RadLex (Radiology Lexicon) \cite{langlotz2006radlex} codes. They accurately summarize all findings in the radiological images together with a detailed attribute description. They are an unstructured collection of findings, with a sequence of annotation terms representing each finding (e.g., "infiltrate/lung/upper lobe/left/patchy/mild"...). The codes use a controlled vocabulary containing 178 terms, which include anatomies, diseases, pathological signs, foreign objects, and attributes. Anatomies and diseases can be matched to broad body regions, such as the respiratory or skeletal system. Attributes include degree, descriptive, and positional attributes. 

We utilize this semi-structured finding representation to construct a highly detailed report template as shown in Figure \ref{fig_template}. Our report template is structured into multiple sections and employs a multi-level hierarchy of questions that delve deeper into the findings at each level. The template can be considered a large decision tree with questions at every level. The highest level asks for the general existence of findings (signs, diseases, abnormal regions, or objects), the second level asks for specific elements, such as a certain object or disease, and the lowest-level questions ask for specific attributes. Table \ref{tab_template-stats} shows how often which question type occurs. To create the template, we parse the codes of all patients and identify all occurrences of term combinations at all levels of the defined hierarchy. We then remove unseen options to produce a streamlined report template that includes only possible options for all findings. Further, we mark all questions as either single- or multi-choice and add a "no selection" option.
\begin{table}[tb]
\centering
\caption{Number of questions and answer options per level in our template. For Level 2 we further list the different topics. *not every answer is an option for all questions}
\begin{tabular}{@{}l|c|c|c|c|c|c|c@{}}
\toprule
                 & L1 & L2  & L2-objects & L2-diseases & L2-signs & L2-abnormal regions & L3  \\ \midrule
Nr. questions & 25 & 216 & 16           & 103           & 65         & 32                    & 477 \\ 
Nr. unique answers & 2 & 2 & 2           & 2           & 2         & 2                    & 94* \\ \bottomrule
\end{tabular}
\label{tab_template-stats}
\end{table}

Overall, our structured report template provides a rigorous and comprehensive framework for classifying radiological images and mimics the style of a structured report in a clinical setting. This enables the development and comparison of methods for the population of structured reports and the prediction of fine-grained radiological findings.

\textbf{Dataset and Evaluation Metrics} Our dataset consists of structured reports for each patient in the IU-XRay data collection, for which finding codes and a frontal X-Ray were available. The new dataset includes 3720 images matched to 3597 structured patient reports entailing more than 180k questions. If multiple images belong to one patient, each image is considered an independent sample. We use a 80-10-10 split to create train, validation and test set. To avoid data leakage, we ensure that different images of the same patient are in the same split.

The goal of our task is to produce fine-grained finding classifications for populating a structured report given an X-Ray image of a patient. This task involves answering a series of questions about the image, gradually adding more detail. We define several evaluation metrics for the proposed benchmark. As the distribution of questions and answers is very imbalanced, we evaluate with the macro precision, recall, and F1 score over all possible paths in the question tree to encourage methods that also perform well in under-represented question-answer combinations. One path is a unique position in the report combined with a specific answer option. Further, we employ report-level accuracy to measure how many predicted reports are entirely correct. During the evaluation, we enforce consistency within the question hierarchy. For example, if the answer to a higher-level question is "no", we prohibit to answer a lower-level question positively. This ensures the generated reports are consistent and coherent, as in a real medical report. Lastly, as multiple instances of an object, sign or pathology can occur for one patient, we iteratively ask for further occurrences, when the model predicts a positive answer. (e.g., "Are there other opacities in the lung?"). We restrict the number of follow-up questions by the maximum of per-patient occurrences in the data. As the order of occurrences is ambiguous, we apply instance matching during the metric computation. We order all predicted instances such that the highest F1 score for this finding is achieved.

\subsection{Hierarchical Visual Question Answering}
With Rad-ReStruct, we propose a hierarchical VQA task, where lower level questions are dependent on context information. For instance, to answer the questions "What is the degree?" it is essential to know what the question is referring to. This information is given through the previous question, which could be "Is there Pneumonia in the lung? Yes". To integrate this context, we propose a hierarchical VQA framework that can effectively answer questions about medical images by considering previously asked questions. We extend the input to the model by pre-pending the current question with the history of previously asked questions and the model's answers. This extension enables interpretable and consistent results. \looseness=-1

\begin{figure}[t]
\includegraphics[width=\textwidth]{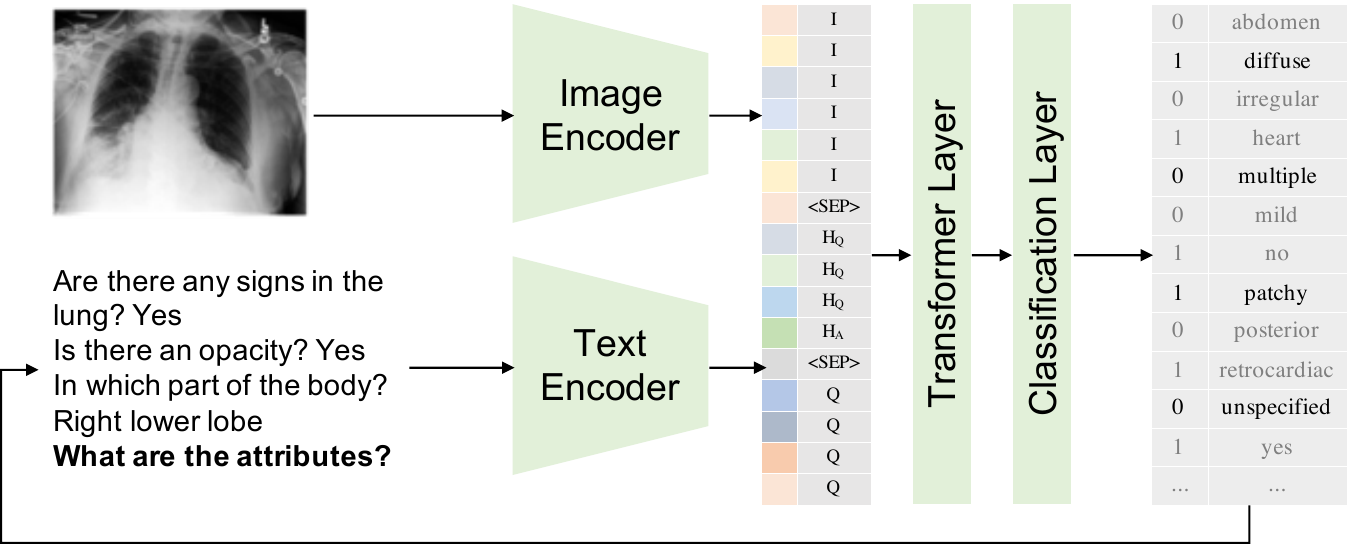}
\caption{Overview of our hierarchical VQA framework. The image and the history with the current question are encoded, concatenated and fused with a transformer layer. The final prediction is computed over the relevant answers and added to the history.}
\centering
\label{fig_model}
\end{figure}

We leverage a pretrained image encoder, EfficientNet-b5 \cite{tan2019efficientnet}, and a domain-specific pretrained text encoder, RadBERT \cite{yan2022radbert}, to extract features from the image, history, and question. The extracted features are fused by a transformer \cite{NIPS2017_3f5ee243} layer, adapted to handle multi-modal input. The fused features are then used to perform a multi-label classification over all answer options. However, we only consider outputs that are valid answers to the current question. For single-choice questions, we predict a single label applying a softmax over all valid answers, while for multi-choice questions, we predict multiple labels using a sigmoid function. Figure \ref{fig_model} shows an overview of the proposed framework.

\textbf{Feature Encoding}  For fusing the image and text features, we construct a token sequence of the form \textit{<image\_tokens> <history\_tokens> <question\_tokens>}. The image tokens consist of the flattened embedding representation of the image encoder, while the history and question text is encoded jointly using RadBERT. The different parts are separated by a <SEP> token and fused by a single transformer layer. We encode the type of input in the token-type IDs with different token types for the image tokens, history questions, history answers, and the current question. Further, we use modified positional encodings to preserve the 2D spatial information of the image as well as the sequence order of the text. We combine 2D positional embeddings \cite{chen2021pix2seq} for the image tokens with the 1D positional encodings used in the original transformer architecture \cite{NIPS2017_3f5ee243} for the text tokens and set the non-used part of the encoding per modality to zero.

\textbf{Training and Evaluation} 
During training, we use teacher forcing, utilizing ground truth history, allowing for efficient batch-wise processing. The model is trained end-to-end, using a weighted masked cross-entropy loss to optimize the classification performance. For every sample in a training batch, we only consider the loss for the labels corresponding to the asked question to avoid optimizing the model on currently irrelevant outputs. The evaluation is autoregressive, thus the model utilizes the previously asked questions and their predicted answers as history. In a hierarchical VQA task such as Rad-ReStruct, the inference is interrupted if the model predicts a negative answer, and sub-questions lower in the hierarchy are automatically answered as negative, enforcing consistency of the prediction. This also improves the explainability of predictions by showing at which level the model made a mistake. For non-hierarchical VQA tasks, the history is utilized solely as context information, allowing the model to exploit inter-dependencies between different questions about the same image.\looseness=-1
\section{Experiments and Results}
We test our model on Rad-ReStruct, setting a baseline for this new dataset. To further validate our model design, we compare the performance of our model with previous work on the standard VQA benchmark VQARad. We train all models on a NVIDIA A40 GPU. We use pytorch-lightning 1.8.3. and the AdamW optimizer with a learning rate of 5e-5 for VQARad and 1e-5 for Rad-ReStruct. For all models, we set the number of epochs by maximizing validation set performance.

\textbf{Rad-ReStruct} For Rad-ReStruct, the history includes all higher-level questions on the same question path. Additionally, attribute questions asked previously about an element, are included in the history, enabling the model to provide consistent predictions. Lastly, the history includes previously predicted instances of the same element. Table \ref{tab:result_radrestruct} and Table \ref{tab:results_rr_detailed} show the overall and question-level results of our model. 
\begin{table}[tb]
\centering
\caption{Results of hi-VQA with and without history and our visual baseline}
\begin{tabular}{@{}l|c|c|c|c|c@{}}
    \toprule
    & domain-specific pretraining data & report acc & F1  & prec & recall \\ \midrule
    Visual baseline & none (only general images) & 31.3 & 30.7 & 65.6 & 31.2 \\ 
    hi-VQA - no history & radiologic reports & 26.2 & \textbf{31.9} & 59.9 & \textbf{34.1}  \\ 
    hi-VQA - RoBERTa\textsubscript{BASE} & none (only general text/images) & 26.2 & 31.6 & 67.9 & 32.4  \\ 
    hi-VQA & radiologic reports & \textbf{32.6} & 31.7 & \textbf{70.7} & 32.1 \\ \bottomrule
\end{tabular}
\label{tab:result_radrestruct}
\end{table}

We compare hi-VQA to a visual baseline, consisting of our image encoder and a classification layer. Hi-VQA achieves better performance than the visual baseline in all metrics, indicating the benefits of targeted information retrieval using a large language model. When comparing the RadBERT text encoder, a RoBERTa model \cite{liu2019roberta} pretrained with radiology reports, to RoBERTa\textsubscript{BASE}, which was pre-trained on general text, the RadBERT encoder is superior. This indicates that our method can benefit from better domain-specific language encoders. Using history information improves report accuracy and precision with a slightly decreased recall and a similar F1 score, showing the benefit of history integration. We emphasize, that the history is especially important for the low-level attribute questions, as these are only meaningful with context. Therefore, it will be even more impactful with improved performance for these questions.
\begin{table}[tb]
\centering
\caption{Detailed performance analysis of our model on Rad-ReStruct. F1, precision and recall are computed as macro average over all paths, where a path is a unique position in the structured report combined with an answer option. The number of answers is the mean count of answer options for all questions belonging to a level.}
\begin{tabular}{@{}l|c|c|c|c|c|c@{}}
    \toprule
    & report acc & F1  & prec & recall & \shortstack{\#paths} & \shortstack{avg \#answers} \\ \midrule
    Level 1 - Topic Existence & 36.6 & 63.8 & 79.0 & 63.5 & 50 & 2\\ 
    Level 2 - Element Existence (all) & 35.7 & 72.2 & 86.0 & 72.3 & 432 & 2 \\ 
    \hspace*{1.07cm} - Diseases & 52.4 & 74.6 & 83.7 & 74.9 & 206 & 2\\ 
    \hspace*{1.07cm} - Signs & 74.3 & 74.1 & 90.1 & 74.1 & 130 & 2\\ 
    \hspace*{1.07cm} - Abnormal body regions & 58.6 & 69.1 & 86.4 & 69.3 & 64 & 2\\ 
    \hspace*{1.07cm} - Objects & 90.4 & 67.8 & 87.6 & 67.1 & 32 & 2\\ 
    Level 3 - Attributes & 32.6 & 3.7 & 60.3 & 4.4 & 1988 & 4.2 \\ \bottomrule
    
    \end{tabular}
\label{tab:results_rr_detailed}
\end{table}

Our labels' hierarchical, structured formulation enables a performance analysis on different topics and levels. Hi-VQA performs well in detecting the existence of sub-topics like objects, diseases, signs, and abnormalities. However, attribute prediction performance is much lower, likely due to the rarity and complexity of these questions and error propagation from higher levels. Such an analysis is precious to understand what a model learned and when it should be trusted.

\textbf{VQARad} is a medical VQA benchmark with 315 radiological images and 3515 questions. The task is to make a classification over 2248 possible answers. In VQARad multiple questions are asked about one image, but in previous work they are always answered separately. To make use of possible inter-dependencies between questions, we define five question levels based on question topics in VQARad, ranging from general to specific: \textit{Modality$\rightarrow$Plane$\rightarrow$Organ$\rightarrow$Presence, Count, Abnormality$\rightarrow$Color, Position, Size, Attributes, Other.} For a certain question, previously asked questions from lower or the same level are included in the history. During training, we augment the history by randomly dropping and reordering questions within a level to prevent overfitting on this small dataset.
\begin{table}[tb]
\centering
\caption{Results of our proposed model hi-VQA on the VQARad benchmark compared to previous work. *RepsNet used an adapted validation in their paper, where unseen answers in the test set are ignored, as they can not be predicted. We calculate their performance when keeping the unseen samples to enable a fair comparison.}
\begin{tabular}{@{}l|l|c@{}}
\toprule
                           & domain-specific pretraining data & acc      \\ \midrule
MEVF \cite{nguyen2019overcoming} & radiologic images  & 66.1  \\
MMQ \cite{do2021multiple}                      & none & 67.0  \\
MM-BERT \cite{khare2021mmbert}                 & radiologic images and reports (joined PT) & 72.0   \\
CRPD  \cite{liu2021contrastive}                 & radiologic images & 72.7  \\
RepsNet   \cite{tanwani2022repsnet}             &  radiologic reports & 73.5* \\
M3AE  \cite{chen2022multi}                 & radiologic images and reports (joined PT) & 77.0  \\
\textbf{hi-VQA - no history} & radiologic reports & 74.5  \\
\textbf{hi-VQA - RoBERTa\textsubscript{BASE}} & none (only general text/ images) & 72.5  \\
\textbf{hi-VQA}             & radiologic reports & 76.3  \\ \bottomrule
\end{tabular}
\label{tab_vqarad_results}
\end{table}

\noindent Table \ref{tab_vqarad_results} shows the performance of hi-VQA compared to previous methods. Amongst the methods without domain-specific joined image-text pretraining, we reach SOTA results, even without history context. When integrating history information, our model achieves competitive results with the current SOTA method, M3AE \cite{chen2022multi}. This result demonstrates the promise of jointly answering questions about the same image in medical VQA tasks. Lastly, we again compare using the RadBERT encoder to RoBERTa\textsubscript{BASE} \cite{liu2019roberta}. We can observe, also on VQARad, using RadBERT improves the performance notably, again indicating that VQA tasks benefit from domain-specific text encoders.
\section{Discussion and Conclusion}
By introducing Rad-ReStruct, the first structured radiology reporting benchmark, we create a framework to develop, evaluate, and compare structured reporting methods. The structured formulation enables an accurate evaluation of clinical correctness at different levels of granularity, focusing on levels with greater clinical importance. Moreover, such a structured finding representation could then again, rule-based, be converted to a text report while maintaining clinical accuracy. To model structured reporting, we present hi-VQA, a novel, hierarchical VQA  framework with a streamlined architecture that leverages history context for multi-question and multi-level tasks. The autoregressive formulation and consistent evaluation increase interpretability and mimic the workflow of structured reporting. Moreover, as each prediction takes previous answers into account, it would allow for an interactive workflow, where the model can make predictions and react to corrections while a radiologist fills out a report.

We set a first baseline for Rad-ReStruct, with particularly good performance on higher-level questions. Although our model has limited performance on the low-level attribute questions, it performed competitive to state-of-the-art on VQARad, indicating the difficulty of our new task. We see this as an opportunity to develop methods for fine-grained understanding of radiology images, rather than solely focusing on higher-level diagnoses. Further, we show the positive effect of history integration, which is crucial for hierarchical and context-dependent tasks such as structured report population. Our work represents a significant step forward in the development of automated structured radiology report population methods, while allowing an accurate and multi-level evaluation of clinical correctness and fostering fine-grained, in-depth radiological image understanding.\looseness=-1

\section*{Acknowledgements}
The authors gratefully acknowledge the financial support by the Federal Ministry of Education and Research of Germany (BMBF) under project DIVA (FKZ 13GW0469C) and the Bavarian Research Foundation (BFS) under project PandeMIC (grant AZ-1429-20C).

%
%
%
\bibliographystyle{splncs04}
\bibliography{refs}

\end{document}